\documentclass[sigconf, screen=true, authorversion=true]{acmart}

\AtBeginDocument{%
  \providecommand\BibTeX{{%
    \normalfont B\kern-0.5em{\scshape i\kern-0.25em b}\kern-0.8em\TeX}}}

\setcopyright{acmcopyright}
\acmDOI{10.1145/3460319.3464811}
\acmYear{2021}
\copyrightyear{2021}
\acmSubmissionID{issta21main-p153-p}
\acmISBN{978-1-4503-8459-9/21/07}
\acmConference[ISSTA '21]{Proceedings of the 30th ACM SIGSOFT International Symposium on Software Testing and Analysis}{July 11--17, 2021}{Virtual, Denmark}
\acmBooktitle{Proceedings of the 30th ACM SIGSOFT International Symposium on Software Testing and Analysis (ISSTA '21), July 11--17, 2021, Virtual, Denmark}

\usepackage{amsmath}

\usepackage{amssymb}
\usepackage{amsfonts}
\usepackage{hyperref}
\usepackage{algorithmic} 
\usepackage{graphicx}
\usepackage{textcomp}
\usepackage{xcolor}
\usepackage{xspace}
\usepackage[ruled,linesnumbered]{algorithm2e}
\usepackage{multirow}
\usepackage{booktabs}
\usepackage{soul}
\usepackage{balance}
\usepackage{tcolorbox}
\usepackage{array}
\usepackage{multirow}
\usepackage{subfigure}
\usepackage{enumitem}

\def\tcc{\leavevmode\rlap{\hbox to \hsize{\color{gray!35}\leaders\hrule height .8\baselineskip depth .5ex\hfill}}}

\newcommand\tool{\textsc{DeepHyperion}\xspace} 
\newcommand\bng{\textsc{BeamNG}\xspace}
\newcommand\mnist{\textsc{MNIST}\xspace}

\newcommand\djan{\textsc{DeepJanus}\xspace}
\newcommand\dlf{\textsc{DLFuzz}\xspace}
\newcommand\mnistts{MNIST\xspace}
\newcommand\beamngts{\textsc{BeamNG}\xspace}

\newboolean{showcomments}
\setboolean{showcomments}{true}         
\ifthenelse{\boolean{showcomments}}
  {\newcommand{\nb}[2]{
  \fbox{\bfseries\sffamily\scriptsize#1}
     {\sf\small$\blacktriangleright$\textit{\textcolor{red}{#2}}$\blacktriangleleft$}
   }
  }
  {\newcommand{\nb}[2]{}
   
  }

\begin{document}

\title[Exploring the Feature Space of DL Systems through Illumination Search]{\tool: Exploring the Feature Space of Deep Learning-Based Systems through Illumination Search}

\author{Tahereh Zohdinasab}
\orcid{0000-0002-0191-1151}
\affiliation{%
  \institution{Universit{\`a} della Svizzera Italiana}
  \city{Lugano}
  \country{Switzerland}
}
 \email{tahereh.zohdinasab@usi.ch}
 
 \author{Vincenzo Riccio}
\orcid{0000-0002-6229-8231}
\affiliation{%
  \institution{Universit{\`a} della Svizzera Italiana}
  \city{Lugano}
  \country{Switzerland}
}
 \email{vincenzo.riccio@usi.ch}

 \author{Alessio Gambi}
\affiliation{%
  \institution{University of Passau}
  \city{Passau}
  \country{Germany}
}
 \email{alessio.gambi@uni-passau.de}

\author{Paolo Tonella}
\orcid{0000-0003-3088-0339}
\affiliation{%
  \institution{Universit{\`a} della Svizzera Italiana}
  \city{Lugano}
  \country{Switzerland}
}
 \email{paolo.tonella@usi.ch}

\newcommand{\changed}[1]{{\color{black}#1}}

\begin{abstract}

Deep Learning (DL) has been successfully applied to a wide range of application domains, including safety-critical ones. Several DL testing approaches have been recently proposed in the literature but none of them aims to assess how different interpretable features of the generated inputs affect the system's behaviour.

In this paper, we resort to Illumination Search to find the highest-performing test cases (i.e., misbehaving and closest to misbehaving), spread across the cells of a map representing the feature space of the system.
We introduce a methodology that guides the users of our approach in the tasks of identifying and quantifying the dimensions of the feature space for a given domain.
We developed \tool, a search-based tool for DL systems that illuminates, i.e., explores at large, the feature space, by providing developers with an interpretable feature map where automatically generated inputs are placed along with information about the exposed behaviours.
\end{abstract}




\maketitle
\pagestyle{plain}


\section{Introduction}
\label{introduction}

Deep Learning (DL) has become an essential component of complex software systems, including autonomous vehicles and medical diagnosis systems. As a consequence, the problem of ensuring the dependability of DL systems is critical.

Unlike traditional software, in which developers explicitly program the system's behaviour, one peculiarity of DL systems is that they mimic the human ability to learn how to perform a task from training examples~\cite{Manning-IIR-2008}. Therefore, it is essential to understand to what extent they can be trusted in response to the diversity of inputs they will process once deployed in the real world, as they could face scenarios that might be not sufficiently represented in the data from which they have learned~\cite{HumbatovaICSE20}.

As discussed in the comprehensive work by Riccio~et~al.~\cite{RiccioEMSE20} and by Zhang~et~al.~\cite{Zhang20}, the Software Engineering research community is working hard at adequately testing the functionality of DL systems by proposing a steadily growing number of approaches. Since part of the program logic of these systems is determined by the training data, traditional code coverage metrics are not effective in determining whether their logic has been adequately exercised. Therefore, recent testing solutions aim at maximising ad hoc white-box adequacy metrics, such as neuron~\cite{PeiCYJ17,GuoJZCS18,TianPSB18,XieISSTA19} or surprise coverage~\cite{KimFY19}, or at exposing misbehaviours~\cite{AbdessalemNBS16,ZhangZZ0K18,GambiMF19}.  A limitation of these approaches is that their output cannot be directly used to explain the behaviour of the DL system under test, e.g. coverage reports do not provide enough information to understand what input features might have caused misbehaviours. Consequently, the usefulness of these approaches for the developers is strongly limited in practice.
\begin{figure}[t!]
 \centering
 \includegraphics[width=0.8\columnwidth]{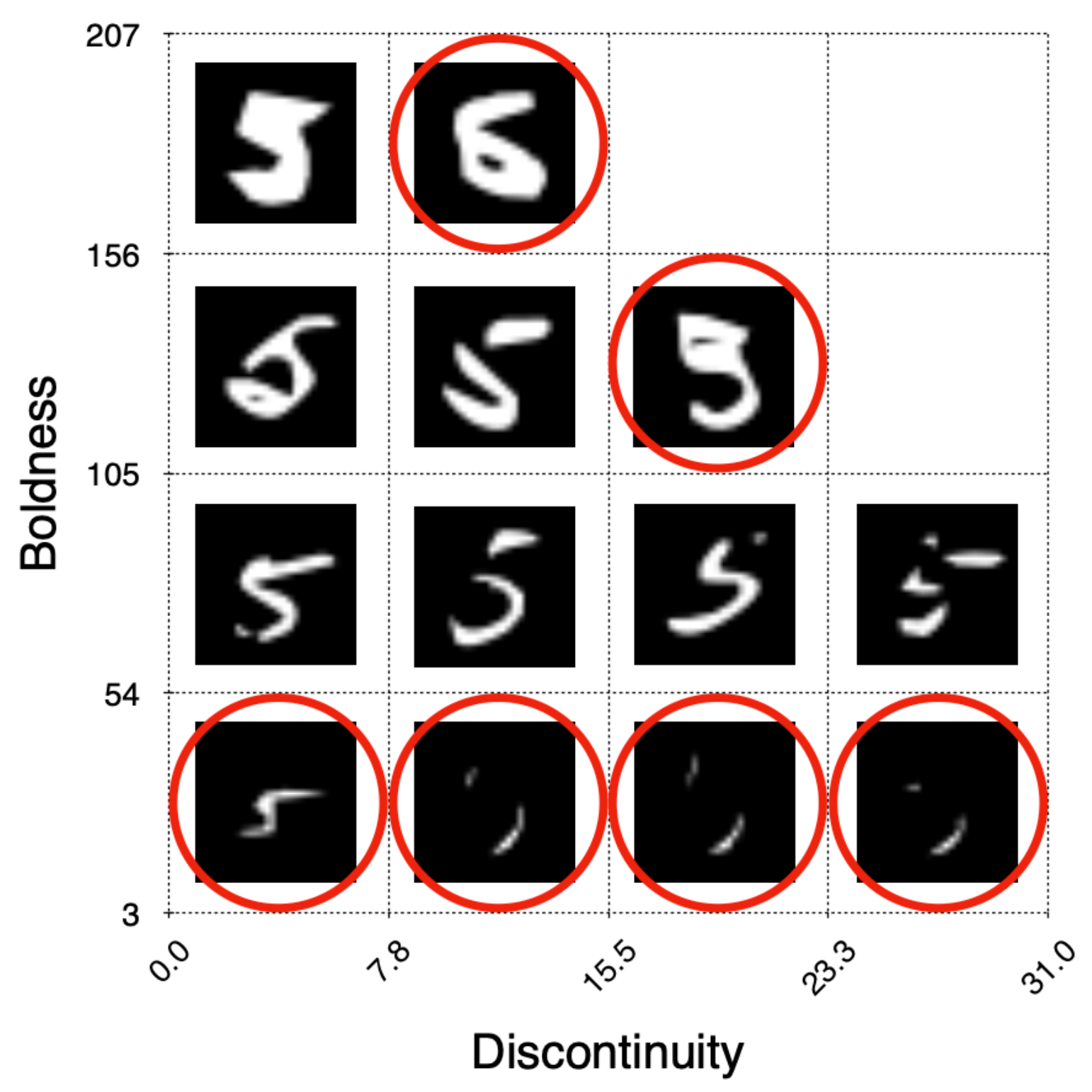}
 \caption{Feature map produced by \tool for a handwritten digit classifier. The two axes quantify two  features: \textit{discontinuity} and \textit{boldness}. Cells show inputs that are either misclassified (marked with a circle) or close to being misclassified.}
 \label{fig:fives}
 \end{figure}
Few approaches~\cite{AbdessalemNBS18,RiccioFSE20} use behavioural properties during test generation, but none of them considers the combination of interpretable features of the DL system under test as the target of test generation. This hinders them from exploring the feature space at large and providing a detailed explanation on how the system behaves for qualitatively different inputs.

In this paper, we introduce a novel way to assess the quality of DL systems by automatically generating a large, diverse set of high-performing (i.e., misbehaving or near-misbehaving), but qualitatively different test inputs that provide developers with a human-interpretable picture of the system's quality. With our approach, developers can understand how different structural and behavioural features of the inputs combine to affect the system's performance. To this aim we developed \tool, an open source automated test input generator for DL systems that leverages the key advantages of Illumination Search, i.e. a family of search algorithms that ``illuminate'' the input space by returning the highest-performing solution at each point of the search space defined by features of interest to the user~\cite{Mouret15}.

\tool is the first approach to apply Illumination Search to DL system testing and to provide  developers with a feature map, where the automatically generated inputs are positioned based on their characteristics and where the misbehaviours they expose can be interpreted (see Figure~\ref{fig:fives} for an example).

A crucial element of our approach is the choice of the dimensions that define the feature space of interest. In particular, the features should represent meaningful properties of the test scenarios, i.e.  discriminative and interpretable properties of the inputs, or behavioural properties manifested by the DL system when exercised by the test inputs. To this aim, we propose a novel systematic methodology that can be used in conjunction with \tool to define the feature dimensions in a domain of interest, making it possible to generate test cases that illuminate the associated map in such domain. This methodology supports the identification of the features that better characterise the generated inputs and the definition of metrics that quantify the selected features. 

We evaluated the proposed technique on both a classification problem (handwritten digit recognition) and a regression problem (steering angle prediction in a self-driving car). Results show that, for both problems, \tool is effective in generating failure-inducing inputs that are structurally or behaviourally different among them, as they cover different regions of the feature space. We compared \tool with state-of-the-art test input generators. Our results show that \tool can explore the feature space at large, whereas existing tools ignore parts of the feature space and expose only misbehaviours that belong to a narrow region of such space.

To foster research and replication, we release the code implementing \tool, the dataset, and all the scripts to replicate the experimental evaluation~\cite{tool}.
\section{The \tool Technique}
\label{technique}

\begin{algorithm}[t]
\caption{\tool's Illumination Search}
\label{algo}

\SetKwInOut{Input}{Input}
\SetKwInOut{Output}{Output}
\Input{$B$: execution budget \\ \textit{seedsize}: seed pool size \\ \textit{popsize}: population size}
\Output{$M$: feature map}

\tcc{/* Initialise the map of elites */}

\textit{map M} $\gets$ $\emptyset$\; 

\tcc{/* Generate initial population */}

\textit{seeds S} $\gets$ \textsc{GenerateSeeds}(\textit{seedsize})\;
\ForEach{$s$ $\in$ S}{
\textsc{Evaluate}(\textit{s})\;
}

\textit{population P} $\gets$ \textsc{InitialisePopulation}($S$, \textit{popsize})\;

\tcc{/* Populate map with initial population */}

\ForEach{\textit{ind} $\in$ P}{
     $M$ $\gets$ \textsc{UpdateMap}(\textit{ind}) \;
}

\tcc{/* Iteratively update the map */}

\While{\textit{elapsedBudget} $<$ $B$} { 

\textit{ind} $\gets$ \textsc{RandomSelection}(M)\;
\textit{ind}$_{\mu}$ $\gets$ \textsc{Mutate}(\textit{ind})\;
\textsc{Evaluate}(\textit{ind}$_{\mu}$)\;
$M$ $\gets$ \textsc{UpdateMap}(\textit{ind}$_{\mu}$)\;
}

\Return(\textit{M})

\end{algorithm}

\tool aims to explore extensively the feature space of a DL system to find the most misbehaving solutions (i.e., those that deviate the most from the expected behaviour) with diverse characteristics. \tool implements the Illumination Search algorithm proposed by J.B. Mouret and J. Clune, named Multi-dimensional Archive of Phenotypic Elites (MAP-Elites)~\cite{Mouret15}. Given \textit{N} dimensions of variation of interest, which define the feature space, \tool looks for the most misbehaving solution at each point in the space defined by those dimensions, trying to fill the entire feature map. The degree of misbehaviour exhibited by a solution is measured by a properly defined \textit{fitness function}. For example, in a grey-scale digit classification problem, two dimensions of interest could be the boldness and discontinuity of the image, whereas the fitness function could be the misclassification probability computed from the softmax layer output of the Deep Neural Network (DNN)~\cite{GoodfellowMIT16}. In such a case, the output of \tool would be a map where each cell is associated with a specific level of boldness and discontinuity, while the entry of each cell would provide an input image with such boldness and discontinuity, which is either misclassified or close to being misclassified -- i.e., corner case inputs with the given features (see~\autoref{fig:fives}).

Algorithm \ref{algo} outlines the top level steps of the Illumination Search implemented in \tool. The map $M$ to be generated is initially empty (line~2). Then a pool $S$ of candidate valid inputs (seeds) is generated (line~4) and evaluated (lines~5-7), which means each seed is associated with its feature values and its fitness function value. The resulting pool of seeds is used to initialise the population $P$ to be evolved by the algorithm (line~8). The position of each initial individual in the map is determined and the highest fitness individual is added to the map in the corresponding position (lines~10-12). Then, the main evolutionary loop is executed, with a termination condition determined by the execution budget (lines~14-19). At each loop iteration, an individual randomly selected from the current map is mutated and evaluated (lines~15-17) and if it has a higher fitness value than the individual in the map cell it occupies, it replaces the existing entry in the map (this is done also if the map entry is currently empty). In the next sections, we describe the key design choices behind \tool's algorithm and how we applied it to the chosen application domains.

\subsection{Model-Based Input Representation} \label{sec:model}

\tool belongs to the family of the model-based input generation techniques~\cite{Utting12}. It does not directly modify raw input data (e.g., pixels) but it manipulates a \textit{model} of the input that is later used to derive the actual raw input data. This enables \tool to generate more realistic inputs, belonging to the input validity domain~\cite{RiccioFSE20}. This implies that \tool is applicable to a given domain if we have a generative model of the input data processed in such domain. 

The development of input models is standard practice in several domains, including safety-critical ones~\cite{Larman1997}. Generative input models are largely domain-specific. In the following, we present two examples of such models for the domains we considered in our experimental evaluation: handwritten digits, processed by a digit classifier, and driving scenarios for self-driving cars.

As regards the handwritten digits, the test inputs are images in the MNIST~\cite{LecunBBH98} format. MNIST is a database of $70\,000$ handwritten digits, originally encoded as 28 x 28 images with greyscale levels that range from 0 to 255. We model them as combinations of B\'ezier curves,  adopting the Scalable Vector Graphics (SVG)\footnote{\url{https://www.w3.org/Graphics/SVG/}} representation. The control parameters that determine the shape of the modelled digit are: the start point, the end point and the control points that define each B\'ezier segment. This representation ensures that the realism of handwritten shapes is preserved even after minor manipulation of the B\'ezier curve parameters~\cite{RiccioFSE20}. We use the Potrace algorithm~\cite{Selinger03} to transform an MNIST input into its SVG model representation. To transform an SVG model back to a $28\times28$  grayscale image, we perform a rasterisation operation by means of the functionalities offered by two popular open source graphic libraries (i.e. LibRsvg\footnote{\url{https://wiki.gnome.org/Projects/LibRsvg}} and Cairo\footnote{\url{https://www.cairographics.org}}).

In the autonomous driving domain, the test input is the scenario in which the car drives. A simulated scenario  can be modelled as the composition of the roads, the driving task (i.e., start point, end point and lane to keep), and the environment, which includes the weather and lightness conditions. We consider input scenarios similar to the those generated by the state-of-the-art testing tool DeepJanus~\cite{RiccioFSE20}. These scenarios consist of plain asphalt roads surrounded by green grass on which the car has to drive keeping the right lane. The environment is set to a clear day without fog. The roads are composed of two lanes with fixed width in which there is a yellow center line plus two white lines that separate each lane from the non-drivable area. Our  model of a road is a sequence of consecutive points in a bi-dimensional space. To produce a smooth and realistic shape for the road being modelled, we use Catmull-Rom cubic splines~\cite{CatmullRom74}. The control parameters that determine the shape of the splines are the coordinates of the control points of the center line spline. To transform the model into a road to be rendered in the simulator, we calculate the road points by means of the recursive algorithm for the evaluation of Catmull-Rom cubic splines proposed by Barry and Goldman~\cite{BarryG88} and the functionality offered by the Shapely library, for the manipulation and analysis of planar geometric objects\footnote{\url{https://github.com/Toblerity/Shapely}}.

\subsection{Fitness Function}

The \textsc{Evaluate} function (lines 6 and 17 in \autoref{algo}) evaluates an individual \textit{ind} by determining the values of its features $\{$\textit{ind}$.f_1$, ..., \textit{ind}$.f_n\}$ and of its fitness function \textit{ind.fitness}, both of which are domain/problem specific. 

For what concerns the definition of the relevant input features in a given domain, we propose a novel, systematic methodology, described in detail in~\autoref{sec:feat-sel}. For what concerns the fitness function, the general idea is that it should quantify how close the DL system is to a misbehaviour. In the following, we illustrate the definition of a sensible fitness function for each of the two domains of handwritten digit recognition and autonomous driving.

For the \textit{digit classification problem}, we exploit the activation levels available in the output softmax layer of the DNN that classifies the input image. In fact, the softmax output can be interpreted as a confidence level assigned to each of the possible classes~\cite{GoodfellowMIT16}, where the selected class for the given input is the one with highest confidence. More specifically, we calculate  the difference between the confidence level associated to the expected class and the maximum confidence level associated to any other class. In this way, we get a fitness value that is close to zero when the correct class and the second highest class have similar activation levels, while we get a negative number when the input is misclassified. Hence, this fitness function is to be minimised.

For the \textit{steering angle prediction problem}, we adopt a fitness function similar to that used by \djan~\cite{RiccioFSE20}. The behaviour of the self-driving system is characterised by the distance of the car from the center of the lane during the simulation~\cite{StoccoGAUSS20, JahangirovaST21}. The fitness  is calculated as $\min (w/2-d)$, where $w$ is the width of the lane and $d$ the distance of the car from the lane centre. The position of the car is approximated by its centre of mass. The fitness function returns its maximum value ($w/2$) when the car is at the center of the lane, whereas it returns a negative number when the car is out of bound. Hence, this fitness function is also to be minimised.

\subsection{Feature Map}

The feature map represents the feature space defined by \textit{N} dimensions of variation that characterise the input or the behaviour of the DL system under test. 
Given the values of an individual's features, \textit{ind}$.f_i, \forall i \in [1:N]$, function \textsc{UpdateMap} (lines 11 and 18 in \autoref{algo})  computes the individual's coordinates in the map $M$ by applying the following mapping function:
\begin{equation}
x_i = \left\lfloor \alpha_i \cdot \mathit{ind}.f_i \right\rfloor
\end{equation} 
The value \textit{ind}$.f_i$ is converted to an integer $x_i$ after multiplying it by a constant $\alpha_i$ that ensures approximately the desired grid size, given the expected range of each feature \textit{ind}$.f_i$. For instance, if the desired grid size is 100 and \textit{ind}$.f_i$ is known to range between 0 and 1, a proper selection of $\alpha_i$ could be $\alpha_i = 100$. The resulting integer $x_i$ is used as an index in the map $M$ to get the cell to be assigned to the individual \textit{ind} (in a 2D map, $M[x_1, x_2]$).

During the search, the size of map $M$ grows dynamically as higher/lower index values $x_i$ are discovered.  In fact, initially $M$ has zero entries along all its dimensions. As soon as a new index $x_i$ is discovered during the search process, $M$ is extended to accommodate the newly discovered range of each dimension. For instance, if the first mapped individual has indexes $(x_1, x_2) = (2, 3)$, the initial map $M$ will have size 1 in both directions and will contain just one cell, at position $(2, 3)$. If later another individual is mapped to $(x_1, x_2) = (5, 1)$, the map $M$ will be extended to cover the integer range [2:5] along its first dimension and [1:3] along the second dimension. Hence, at this point of the search the map will cover the rectangle [2:5] $\times$ [1:3], which means it will contain $12$ cells.

At the end of the search, the final map can be adjusted to allow the user to define a granularity different from the dynamically discovered one. This is particularly useful if users want to compare maps produced in different runs/configurations or by different algorithms. The final remapping is a linear rescaling function:
\begin{equation}
\label{eq:rescaling}
x_i' = \left\lfloor GS \frac{\mathit{ind}.f_i - \mathit{min}_i}{\mathit{max}_i - \mathit{min}_i} \right\rfloor
\end{equation} 
\noindent
where $GS$ is the desired grid size, while $\mathit{min}_i, \mathit{max}_i$ are the minimum/maximum value of the $i$-th feature observed across all maps being rescaled to the new grid.

\subsection{Initial Population Generation}

The generation of the initial population consists of choosing an initial set of diverse individuals from the feature space, given a set of seeds of size \textit{seedsize} and the population size \textit{popsize}. Function \textsc{GenerateSeeds} (line 4 in \autoref{algo}) generates seeds that are valid inputs for the system under test. More specifically, for digit classification, seeds are randomly chosen inputs from the MNIST database and converted to SVG. For steering angle prediction, we randomly generate valid roads.

\tool evaluates the fitness and the feature values of the generated seeds and then it finds the map cells to which they belong (lines 5-7 in \autoref{algo}).
Then, function \textsc{InitialisePopulation} (line 8 in \autoref{algo}) selects as initial population the most diverse inputs among the available seeds, by computing the pairwise Manhattan distance~\cite{Krause86} (sum of the absolute differences of the map coordinates) and greedily constructing the set of most diverse seeds, starting from a randomly selected first seed, up to the desired population size. 

\subsection{Selection and Mutation}

Function \textsc{RandomSelection} (line 15 in \autoref{algo}) randomly chooses an already filled cell in the map~\cite{Mouret15}. The individual that occupies the chosen cell is selected for the next genetic evolution. In this way, \tool is not biased towards the solution with highest fitness, like classic evolutionary algorithms, and it can explore the feature space at large, with the ultimate goal of ``illuminating'' it as completely as possible.

The selected individual is mutated by the \textsc{Mutate} operator (line~16 in \autoref{algo}). This operator manipulates the input model's control parameters by applying a small perturbation to them. \changed{The extent of the perturbation is uniformly sampled in a customisable range.  After applying the operator, \tool verifies that the mutant complies with the constraints of the input domain. For the \textit{digit classification} problem, the mutated control points must remain within the 28 x 28 input grid. For the \textit{steering angle prediction} problem, \tool enforces the following constraints to ensure that the mutant is a valid road: (1) the start point and the end point of the road should be different, (2) the road should fall within a square bounding box of fixed size, and (3) a road should not self-intersect.} Moreover, \tool also verifies that, once concretised into an actual input for the DL system, the mutant is different from its parent. If any of these checks fails, the operator is applied repeatedly, until a valid input is obtained.
\section{Definition of the Map Dimensions} \label{sec:feat-sel}

\begin{figure}[t!]
 \centering
 \includegraphics[width=1.0\columnwidth]{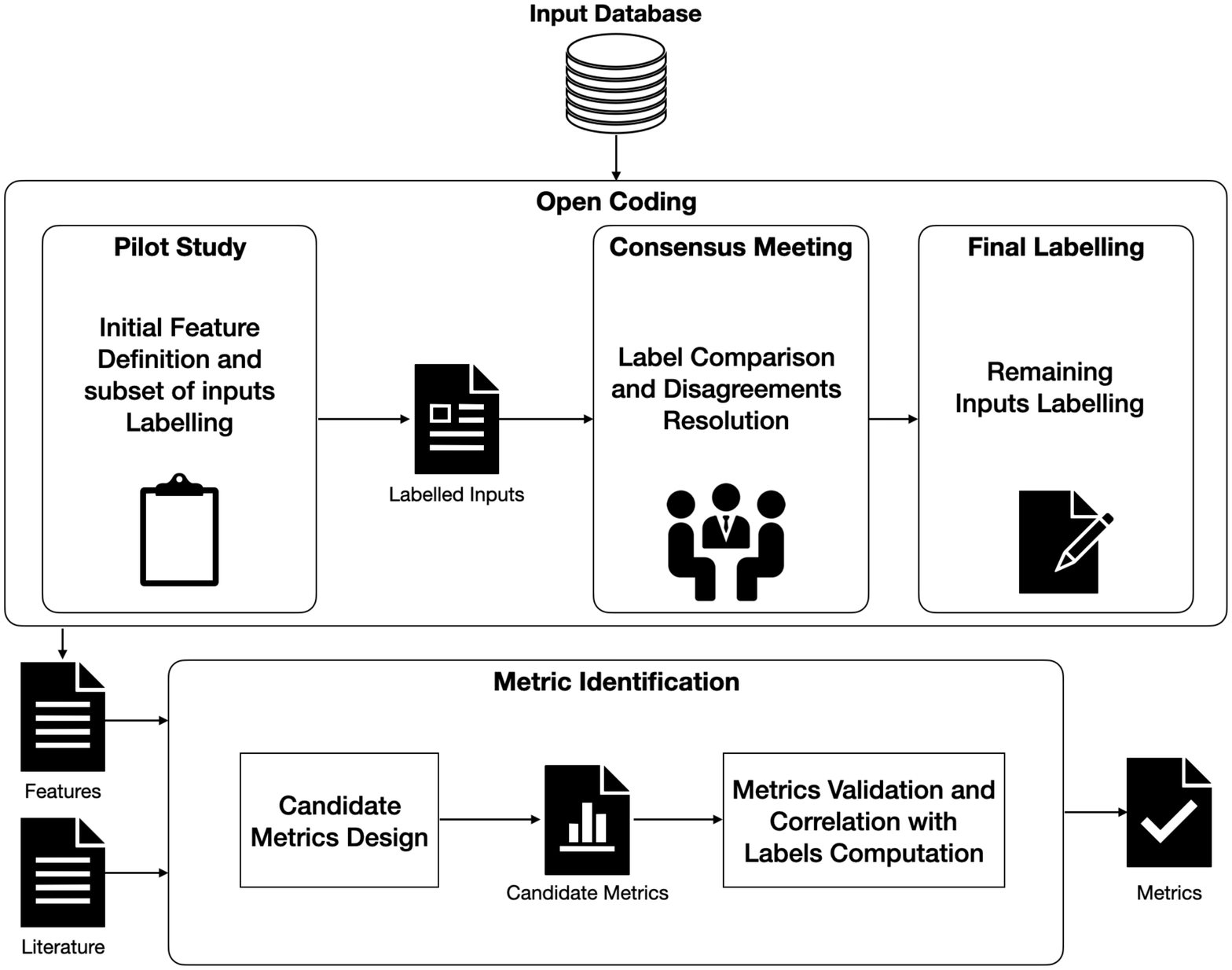}
 \caption{Feature Selection Methodology}
 \label{fig:feat}
 \end{figure}

A crucial element of our approach is the choice of the dimensions of variation of the automatically generated test cases. Such dimensions define both the feature space of interest to the user~\cite{Mouret15} and the search space of \tool. In the case of DL testing, they should represent meaningful properties of the test scenarios: either discriminative and interpretable \textit{structural features} of the inputs, or \textit{behavioural features} observed as the DL system processes the input and produces its output.

We propose an empirical methodology that can be used to define the feature dimensions in a new domain of interest. 
Our methodology consists of two macro-steps (see Figure~\ref{fig:feat}): (1) \textit{open coding}: select the features that better characterise the generated inputs, and (2) \textit{metric identification}: quantify the selected features. The second step is  needed to provide \tool with quantitative feature values to position the generated tests in the feature map.

\changed{This methodology relies on the experts' ability to define meaningful features and metrics to quantify them. Therefore, it can be challenging for DL systems with complex input/output spaces.}

\begin{table}
\setlength{\tabcolsep}{3pt}
\renewcommand{\arraystretch}{1.1}
\centering
\caption{Correlation between features and metrics}
\begin{tabular}{ l l l r r r}

\toprule

 Case Study & Feature & Metric & Agree & Correl & $p$-val\\ 
 
 \midrule
 
 \multirow{4}{*}{MNIST} & Boldness & Lum & 100\% & 0.67 & <0.002 \\  
 & Smoothness & AvgAng & 66\% & 0.05 & 0.241 \\
 & Discontinuity & Mov & 100\% & 0.90 & <0.002 \\
 & Rotation & Or & - & 0.43 & <0.002  \\
 
 \bottomrule

 \multirow{3}{*}{BeamNG} & Smoothness & MinRad & 95.8\% & 0.58 & <0.002  \\  
 & Complexity & TurCnt & 87.5\% & 0.63 & <0.002 \\
 & Orientation & DirCov & 89.5\% & 0.66 & <0.002 \\

 \bottomrule

\end{tabular}

\label{table:correlation}
\end{table}

\subsection{Open Coding}
The first step entails an \textit{open coding procedure}~\cite{Seaman99} in which a set of existing inputs is manually analysed by human assessors to select the relevant features in a given domain. 
Since we are interested in both structural and behavioural features, the information provided to the human assessors is not restricted to the bare inputs (i.e. digit images and roads):  it also includes the output of the DL system when processing the given existing inputs (e.g., the class predicted by an image classifier), as well as any relevant behavioural data (e.g., the trajectory of the car driving on the input road).

The assessors independently tag the inputs assigned to them by either reusing an existing feature label or defining a new one. Each feature label is composed of a feature name, paired with the corresponding feature value, chosen from a rating scale, usually with five levels. For instance, a hypothetical \textit{speed of a self-driving car} label will have values that range between -2 and +2, where -2 means ``very low'', while +2 means ``very high''. 
This procedure is supported by a web application that we developed, which ensures that unlabelled inputs are equally distributed among the assessors, enables assessors to label inputs according to the existing features as well as to define new features, and supports conflict resolution when assessors evaluate the same input differently.

In our methodology, it is strongly advised to run a preliminary pilot study on a subset of inputs to gain confidence in the labelling procedure and, more importantly, agree on the meaning of the features and on the interpretation of the corresponding values. The pilot is concluded with a consensus meeting in which the disagreements are solved either by consensus among the assignees or arbitration by the other assessors. In our experience, a disagreement is worth being discussed in the consensus meeting when the  assigned  values differ by more than 1 position in the rating scale (e.g., a disagreement between ``very low'' and ``low'' speed can be just ignored, while one between ``low'' and ``high'' is worth being discussed and solved). It might happen that the assessors realise through the discussion that some important features have been overlooked. Therefore, as part of the consensus meeting, assessors are allowed to agree upon additional features to be considered during the labelling procedure.

Only when a common understanding of the features and of their possible values is reached, we suggest that it is possible to switch from the pilot study mode to the final study mode. In the final study, it is usually enough that each remaining unlabelled input is evaluated by a single assessor. In fact, while during the pilot study the number of inputs being labeled is kept small, in the final labelling phase we typically want to label as many inputs as possible.

\subsection{Metric Identification}

\changed{The second step of our methodology aims to define a set of metrics that can accurately quantify the domain-relevant features. The metrics can be either (1) selected from the most used in the literature or (2) designed ad-hoc to accurately quantify the features identified in the Open Coding step.}
 
To select the most accurate metrics for the features \changed{that have been identified in the previous step}, we compute the Pearson correlation coefficient~\cite{Benesty2009} and the associated $p$-value, between the manually defined feature values, converted from the rating scale to a numeric scale (e.g., in the range [1:5]), and the values returned by the candidate metrics. The metrics with highest, statistically significant ($p$-value $<$ 0.05) correlation are chosen to quantify the selected features. 
In the following, we provide the details about how this methodology was applied to each of our case studies, i.e. digit recognition and autonomous driving.

\subsection{Dimensions for Digit Recognition} \label{sec:MNIST-feat}

\subsubsection{Open Coding}

In this phase, three authors acted as assessors. In the pilot, we randomly selected $30$ images from the MNIST database and each assessor was assigned $20$ images, such that each input was evaluated by two assessors. The assessors identified the following features, to which they assigned values within a range from -2 to 2:
\begin{itemize}[leftmargin=*]
	\item \textbf{Boldness}, indicates how strong the stroke of the handwriting is. It ranges from very thin ($-2$) to very thick line ($2$).
	\item \textbf{Smoothness}, indicates the absence of sharp angles in the digit. It ranges from sharp angles ($-2$) to smooth angles ($2$).
	\item \textbf{Discontinuity}, indicates how continuous the stroke of the handwriting is. It ranges from continuous line ($-2$) to digits made of multiple disconnected segments ($2$).
	\item \textbf{Rotation} with respect to the vertical axis. It ranges from strongly tilted to the left ($-2$) to strongly tilted to the right ($2$). 
\end{itemize}
Examples of images of handwritten digits at various levels of Boldness and Discontinuity can be found in Figure~\ref{fig:fives}. The inter-rater agreement during the pilot study, measured as the percentage of inputs that were assigned the same feature value or feature values with a difference of 1, is reported in  \autoref{table:correlation} under \textit{Agree}. We observed that assessors strongly agreed over Boldness and Discontinuity (i.e., no conflicts have been registered). Noticeably, \autoref{table:correlation} does not report any agreement value for Rotation, as the assessors introduced this feature during the consensus meeting, i.e., after the data collection for the pilot study ended.
In the final phase, we randomly selected $600$ images from MNIST and each of the three assessors labelled $200$ images.

\subsubsection{Metric Identification}

To measure each feature resulting from the labelling procedure, we designed several candidate metrics and applied them to the $630$ images labelled by the assessors. ~\autoref{table:correlation} (top) shows the metric with highest correlation for each MNIST feature, together with the corresponding correlation and $p$-value:
\begin{itemize}[leftmargin=*]
	\item \textbf{Luminosity} (Lum): number of light pixels of the image, i.e.,  pixels whose value is  above $127$. 
	\item \textbf{Average Angle} (AvgAng) the average angle of the Bezier curves in the SVG representation of the digit.
	\item \textbf{Moves} (Mov): sum of the Euclidean distances between pairs of consecutive sections of the digit. To obtain the sections of a digit, we convert its bitmap to SVG.
	\item \textbf{Orientation} (Or): vertical orientation of the digit, obtained by computing the angular coefficient of the linear regression of the non-black pixels, i.e., pixels with value greater than $0$. 
\end{itemize}
As shown in \autoref{table:correlation}, for Boldness, Discontinuity and Rotation we were able to define metrics that significantly correlate with the human assessment, whereas this was not possible for Smoothness, which turned out to be both difficult to evaluate for humans (see low inter-rater agreement) and difficult to quantify precisely. Hence, this feature was not included among those used by \tool for input generation.

\begin{figure}[t!]
 \centering
 \includegraphics[width=0.8\columnwidth,]{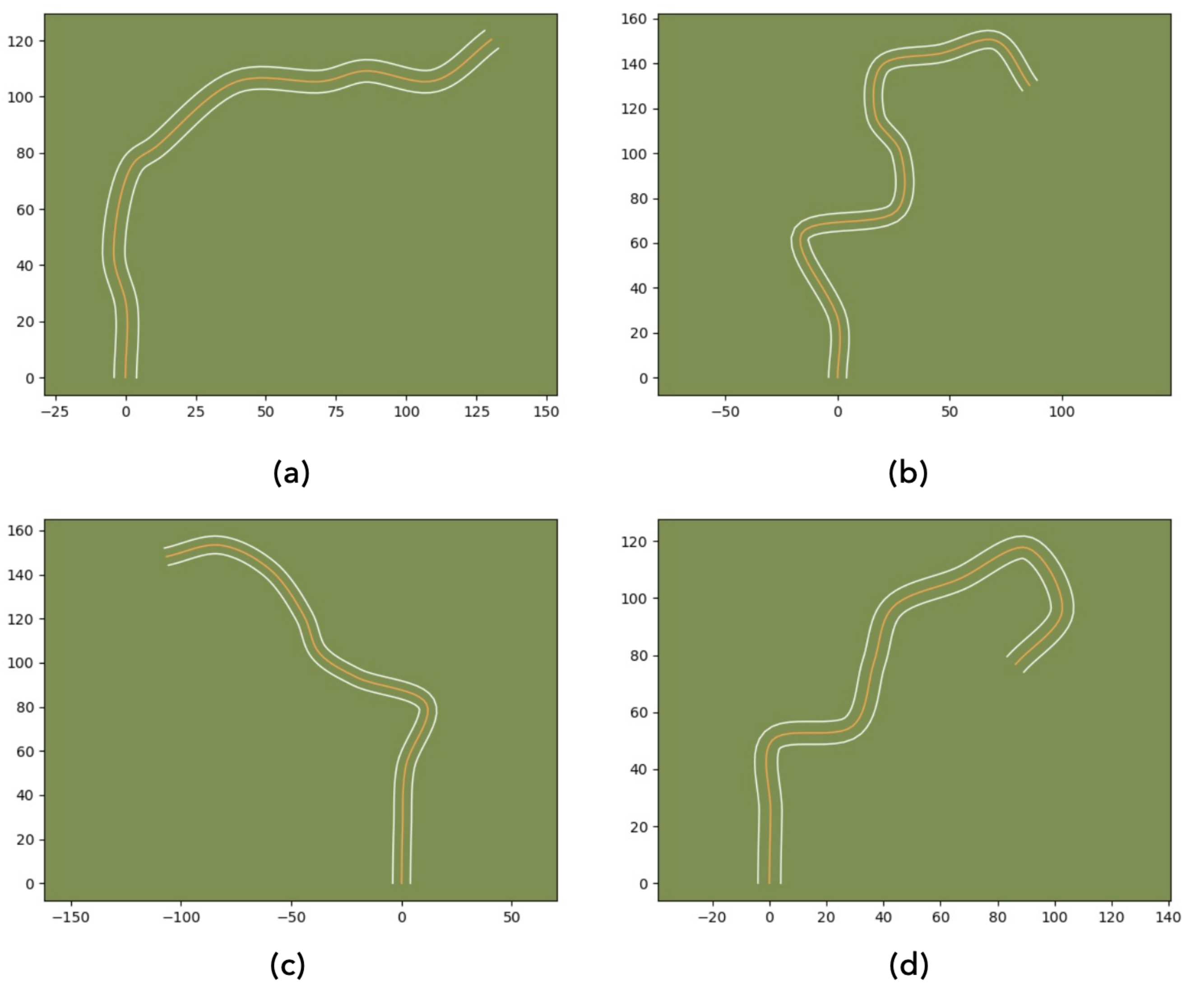}
 \caption{Examples of virtual roads}
 \label{fig:roads}
 \end{figure}

\subsection{Dimensions for Autonomous Driving} \label{sec:BNG-feat}

\subsubsection{Open Coding}

In this phase, all the authors acted as assessors. In the pilot, we randomly generated $40$ virtual roads according to our model representation.
Each assessor was assigned $20$ images representing roads, so that each road was evaluated by two assessors. To simplify the job of the assessors, the web application supporting the labelling procedure provides some interaction facilities for the inspection of the road, such as: (1) zoom in/out; (2) selection of specific road segments; (3) navigation along the road; (4) toggling the visualisation of the car.
The images abstract the  roads over a two-dimensional plane but retain their geometrical properties and the proportions to the vehicle. In the images we draw boxes that represent the vehicle and cones that represent its field of view, to give assessors more context.

The assessors identified the following  features, to which they assigned values within a range from $0$ to $5$:
\begin{itemize}[leftmargin=*]
	\item \textbf{Smoothness}, indicates how smooth the turns of the road are. It ranges from sharp turns ($0$) to gentle turns ($5$).
	\item \textbf{Complexity}, indicates how complex the road's shape is. It ranges from almost straight roads ($0$) to roads with many turns ($5$).
	\item \textbf{Orientation}, indicates how many directions (i.e., N, NE, E, SE, S, SW, W, NW) the road covers. It ranges from straight road which is oriented to one direction only ($0$) to road that covers the whole spectrum of directions ($5$).
\end{itemize}
\changed{\autoref{fig:roads}.a reports a smooth but complex road, whereas the road in \autoref{fig:roads}.b is complex and has sharper turns. The road in \autoref{fig:roads}.c is smooth but includes a very sharp turn. \autoref{fig:roads}.d shows a road covering almost the whole spectrum of directions.}

As reported at the bottom of \autoref{table:correlation}, during the pilot, we observed that assessors generally agreed upon all the features. In the final phase, we randomly generated $400$ roads and each assessor tagged $100$ of them.

\subsubsection{Metric Identification}
We designed a set of candidate metrics and applied them to the $440$ images labelled by the assessors. 
We eventually selected the following $3$ metrics that best correlate with the corresponding features, as reported in ~\autoref{table:correlation} (bottom):
\begin{itemize}[leftmargin=*]
	\item \textbf{Minimum radius of curvature} (MinRad): minimum value of the radius for the circles passing through triplets of consecutive road waypoints.
	\item \textbf{Turn Count} (TurCnt): number of turns in the road, where a turn is a change of direction between consecutive road segments by more than $5^{\circ}$.
	\item \textbf{Direction Coverage} (DirCov): number of different angular sectors covered by the directions of the road segment. In particular, we consider $36$ sectors, each spanning $10^{\circ}$.
\end{itemize}
In addition to the features that characterise the structure of the test input, we considered further features to capture the behaviour of the car during the simulation. In particular, we used the following metrics that have been proposed as quality metrics for self-driving cars~\cite{JahangirovaST21} to measure the \textit{quality of driving}:
\begin{itemize}[leftmargin=*]
\item \textbf{Standard deviation of the steering angle} (StdSA): standard deviation of the sequence of steering angles collected along the road during self-driving.
\item \textbf{Mean lateral position of the car} (MLP): mean distance between the center of the car and the center of the driving lane, where the mean is computed across all car positions observed along the road.
\end{itemize}
\section{Experimental Evaluation}
\label{experiment}

\subsection{Subject Systems}

We evaluate \tool on two DL systems which address different tasks and belong to different domains. 
Moreover, they have been widely used in the literature to evaluate testing techniques for DL systems~\cite{RiccioEMSE20, Zhang20}. Hereafter, we refer to these systems as \mnist and \bng, respectively.

The \mnist system performs a classification task, which consists of recognising handwritten digits from the MNIST dataset~\cite{LecunBBH98}. It is a DNN that predicts which digit is represented in a greyscale image. We considered the DNN instance provided by Keras,~\cite{convnet} because of its popularity. It has $99.8$\% test accuracy, obtained after training it ourselves on the MNIST training set with its default configuration, i.e. $12$ epochs, batches of size $128$, and a learning rate equal to $1$.

The \bng system is a self-driving car equipped with a Lane Keeping Assist System (LKAS). The DL component solves a regression problem, i.e., it predicts the steering angle of the car given the image of its onboard cameras. We tested the whole DL system which includes the LKAS by using the BeamNG.research driving simulator~\cite{beamng_research}, a freely available research-oriented version of the commercial game BeamNG.drive. 
The DL component driving the car utilises the \textsc{DAVE-2} architecture designed by Bojarski et al. at \textsc{NVIDIA}~\cite{BojarskiNVIDIA16}, consisting of three convolutional layers, followed by five fully-connected layers. The DNN was trained with images captured by the camera sensors of the car, paired with the steering angles provided by the simulator's autopilot, which takes advantage of global knowledge and computes the optimal steering angle geometrically.  We trained the model for $4,600$ epochs, with batches of size $128$ and a learning rate equal to $0.001$. We used a training dataset obtained by letting the autopilot drive up to 15 mph on $30$ randomly generated roads.

\subsection{Research Questions}

The goal of our evaluation is to understand whether coupling feature maps and automated test generation is an effective technique for DL testing, which (1) can thoroughly stress the DL system under different conditions, and (2)  can provide information useful to characterise problems in DL systems. Therefore, we seek to answer the following research questions:

\noindent\textbf{RQ1 (Failure Diversity):}  \emph{How effective is \tool in generating test inputs that expose diverse failures?}

Generating tests that trigger failures is more useful when these failures are diverse. Whereas, a test generator that repeatedly exposes the same problem is not desirable, as it wastes computational resources.

\textbf{Metrics:} To assess how many different failures are triggered during a run, we measure the number of \textit{Mapped Misbehaviours (MM)}, i.e. how many cells of the feature map M contain at least one failure-inducing input.

To measure how the mapped misbehaviours of M are diverse, we compute the \textit{Misbehaviour Sparseness (MS)}, defined as the average maximum Manhattan distance between cells containing misbehaviours ($MM \subseteq M$):
\begin{equation}
\label{eq:MS}
\text{MS}(M) = \frac{\sum_{i \in MM} \max_{j \in MM} \text{dist}(i, j)}  {|MM|}
\end{equation}
\noindent\textbf{RQ2 (Search Exploration)}: \emph{How extensively does \tool explore the feature space?}

Effective test generation should exercise different behaviours of the systems under test. This can be achieved by exploring the feature space extensively, at large.

\textbf{Metrics:} We measure the map coverage as the number of \textit{Filled Cells} in the map ($FC \subseteq M$).
Moreover, to measure how broadly our tool explores the feature space, generating inputs in diverse cells, we use the \textit{Coverage Sparseness (CS)}, defined as the average maximum Manhattan distance between filled cells:
\begin{equation}
\label{eq:FS}
\text{CS}(M) = \frac{\sum_{i \in FC} \max_{j \in FC} \text{dist}(i, j)}  {|FC|}
\end{equation}
\noindent\textbf{RQ3 (Feature Discrimination)}: \emph{How strongly do different combinations of features characterise the failure-inducing inputs?}

The existence of regions of the map where the probability of misbehaviours is very high indicates that the corresponding feature value combinations are very likely to induce a failure, since most of the times when the combination was generated by \tool, a misbehaviour was observed. This could provide developers with a powerful tool to understand the conditions responsible for misbehaviours (a form of root-cause analysis).

\textbf{Metrics:} To answer this research question, we compute the \textit{Misbehaviour Probability (MP)} for each cell of a map as the ratio between the number of failure-inducing inputs and the total number of inputs generated by \tool during the search process for that cell. Since occasionally only a small number of inputs may be generated by \tool for a given cell during the search, our estimate of MP might be affected by a large error. Hence, we also compute the confidence interval of MP. In particular, we use Wilson's confidence interval estimator for binomial random variables: in our case, such binomial variable indicates whether a misbehaviour is induced or not. We consider a combination of feature values with a high probability of failure if its MP value is greater than $0.8$ and the lower bound of its confidence interval is above $0.65$.

\subsection{Experimental Procedure}

\begin{table}
 \centering
 \caption{\tool Configurations}
 \begin{tabular}{l r r} 
 \hline
 Parameter & MNIST & BeamNG \\ 
 \hline
 seed pool size &  900  & 40\\
 population size &  800   &  24  \\
 time budget (s) &  3600 &   36000   \\
 mutation lower bound  & 0.01 & 1  \\
 mutation upper bound &  0.6  &  6 \\ 
 \hline
 \end{tabular}
 \label{table:configurations}
 \end{table}

To answer our research questions, we ran \tool with different combinations of the features we identified following our methodology (see~\autoref{sec:feat-sel}). We limited \tool to use only pairwise combinations of features to ease the visualisation of the maps and the discussion of the results; however, the algorithm is general and works also with maps that have more than two dimensions. For \mnist, we report the results achieved by considering all the three features that significantly correlate with the corresponding metrics (see \autoref{table:correlation}). As regards \bng, we conducted our experiments on five feature combinations that cover the three combinations types: two structural features, two behavioural features, and a combination of a structural and a behavioural feature.

We compared \tool with state-of-the-art testing approaches for DL systems, i.e. \djan~\cite{RiccioFSE20} and \dlf~\cite{GuoJZCS18}.

\dlf is representative of approaches that generate test inputs for image classifiers by applying perturbations to the raw input (i.e., pixels), often used to generate adversarial examples and test the robustness of DL components. It has been applied to the \mnist system. However, it cannot be applied to \bng since it could only manipulate  individual camera inputs, without affecting the road shape.

\djan is a search-based tool that generates inputs at the frontier of behaviours of DL systems, i.e. similar inputs that trigger different system behaviours. It is a model-based approach that can be applied to both \bng and \mnist systems. Moreover, it shares with \tool the same input representation which guarantees a consistent measurement of the features and, thus, a fair comparison of the approaches. 

We ran each tool the same number of times on each subject system, i.e. $30$ times on \mnist and $10$ times on \bng, respectively. To ensure a fair comparison, we ran them on the same computing nodes and used the same time budget for each tool, i.e. $3600$ seconds for \mnist and $36000$ seconds for \bng, respectively. 
The reason for the different time budgets is that testing \mnist requires only to feed it an image and get the corresponding prediction, which usually is a matter of milliseconds, while testing \bng requires to execute driving simulations that take several minutes to complete.

The configurations of \tool were obtained in a few preliminary runs and are reported in \autoref{table:configurations}. With the other tools, we either used the configuration reported as the one achieving the best performance or directly contacted the tools' corresponding authors when some details about the configuration were missing.

The initial seeds for \mnist were obtained by randomly selecting $900$ inputs from the MNIST test set, all belonging to the class ``5''. We obtained similar results for other digit classes, but we do not report them for space reasons. For \bng, the seeds were defined by 10 control points in which the initial point was always at a fixed position, whereas the others were placed at a random position 25 meters away from the previous one.
 
At the end of the runs, we used the inputs generated by each tool and the outputs generated by the subjects to compute the feature map of each run. All the maps were generated with the same number of cells for each feature, i.e. up to $25$ cells, by using the rescaling function described in~\autoref{eq:rescaling}. The min/max values defining the range for each feature were the ones observed across the runs of all the tools. We used these feature maps to compute the metrics associated with each research question. 
To assess the statistical significance of the comparisons between \tool and the considered state-of-the-art tools, we performed the Mann-Whitney U-test and measured the effect size  by means of the Vargha-Delaney's \^A$_{12}$ statistic~\cite{Arcuri14}.
\section{Results}
\label{results}
 
\subsection{RQ1: Failure Diversity}

\begin{figure}
\centering
	\includegraphics[width=\columnwidth]{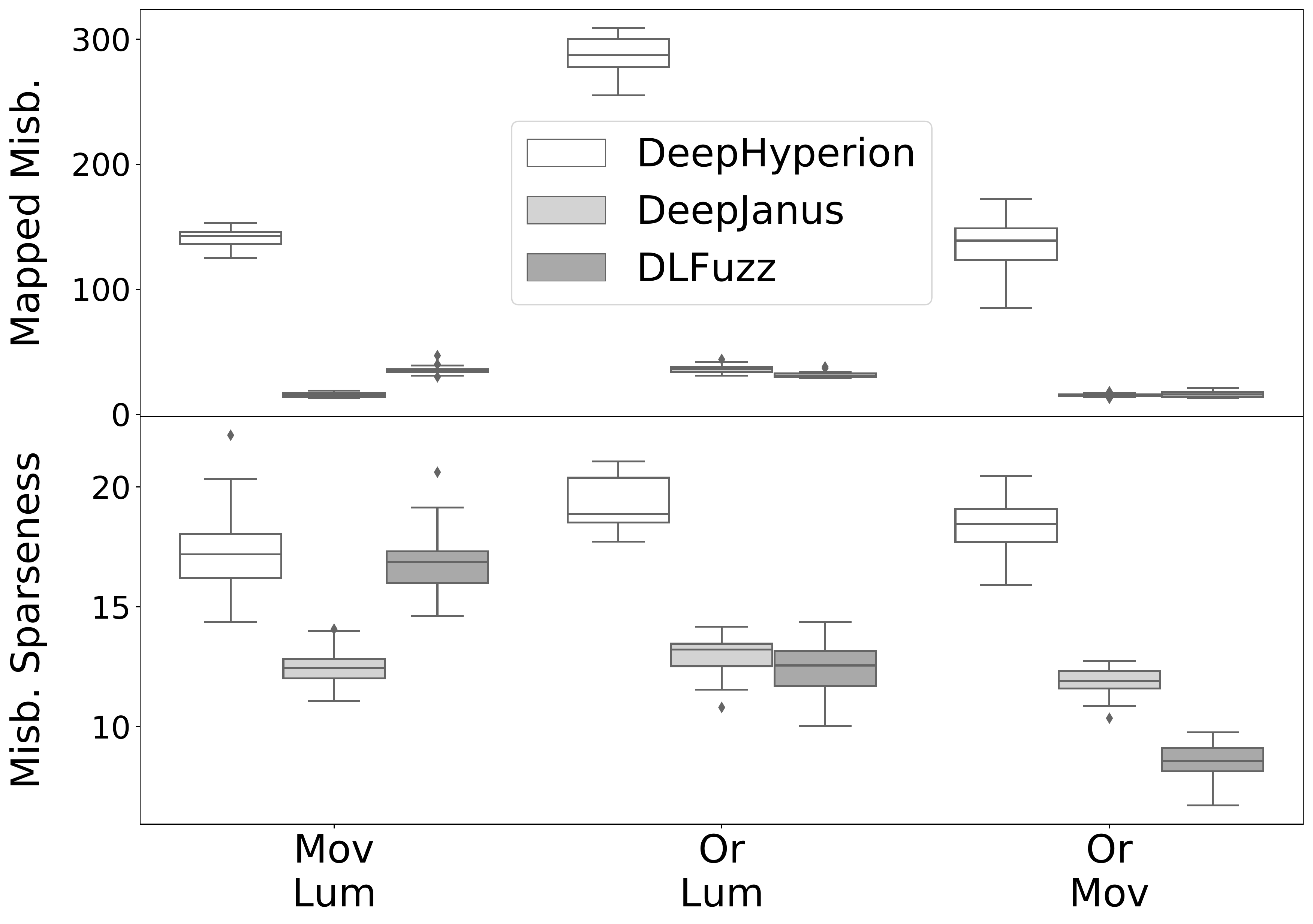}
        \caption{RQ1: Misbehaviours found by \tool, \djan and \dlf on \mnist}
        	\label{fig:rq1-MNIST}
\end{figure}
	
\begin{figure}
	\includegraphics[width=\columnwidth]{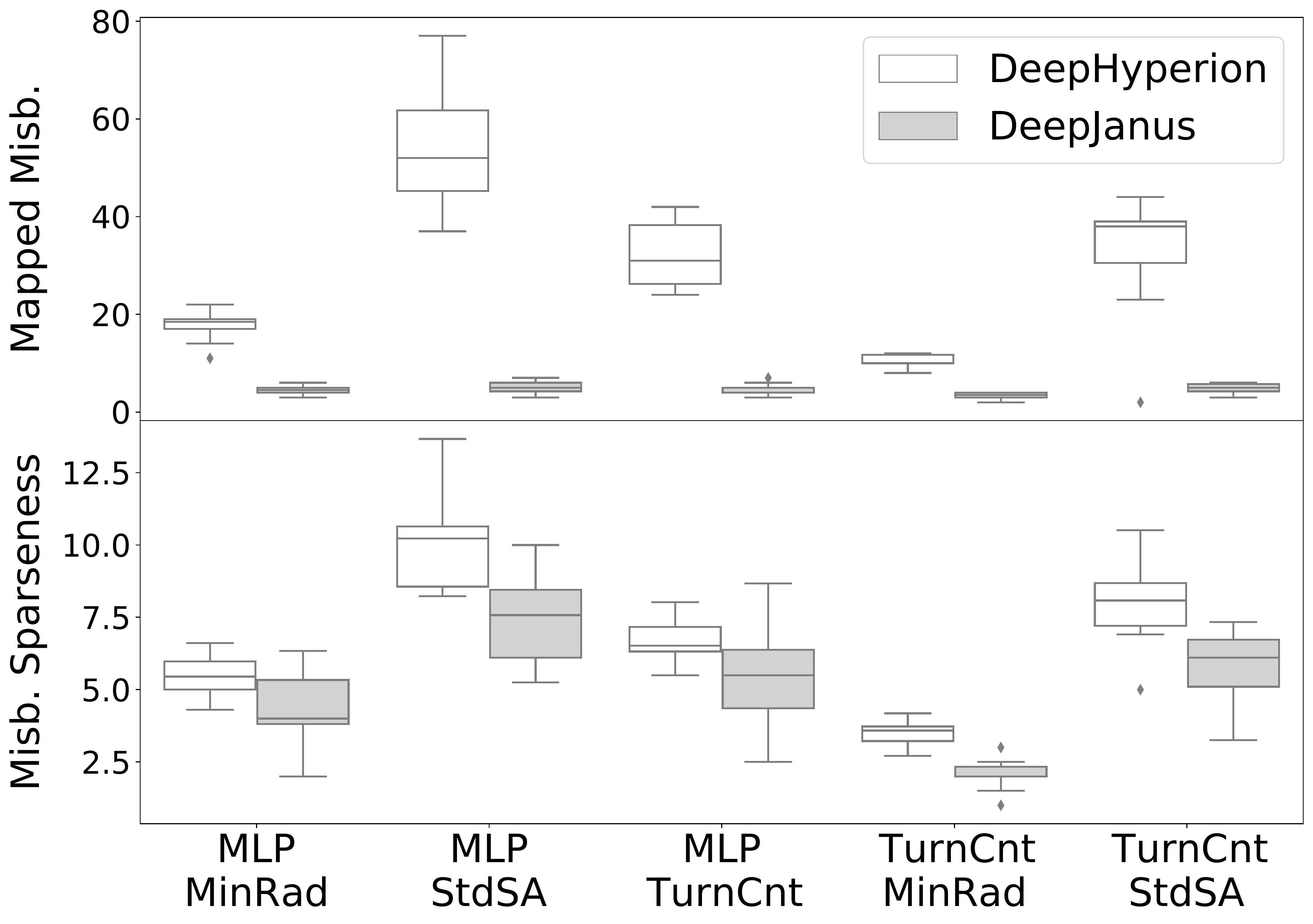}
        \caption{RQ1: Misbehaviours found by \tool and \djan on \bng}
	\label{fig:rq1-BeamNG}
\end{figure}

~\autoref{fig:rq1-MNIST} shows the number of diverse misbehaviours found by each tool in the \mnist system and their sparseness on the map. 

As shown in ~\autoref{fig:rq1-MNIST} (top), \tool found more than \changed{$100$} diverse misbehaviours for each feature combination. \tool outperformed the other tools by generating a higher number of diverse failures for all feature combinations ($p$-values  $<0.05$; large effect size). In particular, for the \textsc{Or-Lum} feature combination \tool produced a number of mapped misbehaviours that is remarkably above all the other tools (exceeding the second best by more than $250$ misbehaviours). 

~\autoref{fig:rq1-MNIST} (bottom) shows that \tool produced failure-inducing inputs that are more sparse on the feature map for all feature combinations, as its Misbehaviour Sparseness (MS) metric is always significantly higher than the compared tools with $p$-values $<0.05$ (effect size is always large with the exception of \textsc{Mov-Lum} vs \dlf, for which the effect size is small). This result is achieved despite \djan explicitly rewards diversity, having a fitness function that promotes the euclidean distance among solutions. \tool can expose a large number of misbehaviours (\autoref{fig:rq1-MNIST}, top), and, more importantly, it can reveal highly \textit{diverse misbehaviours}, associated with very distant feature combinations (\autoref{fig:rq1-MNIST}, bottom).

\autoref{fig:rq1-BeamNG} shows Mapped Misbehaviours and Misbehaviour Sparseness of the tools that have been applied to the \bng system.
\autoref{fig:rq1-BeamNG} (top) shows that \tool was always able to expose several diverse failures of the \bng system (at least $10.5$ on average). In comparison with \djan, it produced significantly more misbehaviours for all the feature combinations ($p$-values $<0.05$, large effect size). In particular, \textsc{MLP-StdSA} produced an impressive number of mapped misbehaviours ($53.4$ on average) which is remarkably above the competitor (exceeding it by $48.3$). The goodness of the combination of behavioural features is confirmed also on the sparseness of the misbehaviours, as shown in \autoref{fig:rq1-BeamNG} (bottom), since it performed better than the other combinations, i.e. $10.14$ on average. With respect to the competitor, \tool generated significantly sparser inputs for four out of five combinations. Only for \textsc{MLP-TurnCnt}, the sparseness values of the inputs generated by the two tools do not show any statistically significant difference ($p$-value $= 0.06$, slightly above the conventional threshold of 0.05; medium effect size in favour of \tool). 

We also compared the total number of misbehaviours exposed by each tool, regardless of their diversity. \autoref{table:totalmisb} shows that \tool exposed a total number of misbehaviours significantly larger than the competitors ($p$-values $<0.05$, large effect size). 

This confirms that the good results achieved by \tool are not biased by the size of the feature maps we adopted in the experiments \\

\begin{table}
\setlength{\tabcolsep}{3pt}
\renewcommand{\arraystretch}{1.1}
\centering
\caption{Total number of misbehaviours found by \tool, \dlf and \djan on \mnist and \bng}
\begin{tabular}{ l l r}

\toprule

 Subject & Tool & Misbehaviours\\ 
 
 \midrule
 
 \multirow{5}{*}{MNIST} & DH-Mov-Lum & 2657.83 $\pm$ 441.02 \\  
 & DH-Or-Lum & 7292.1 $\pm$ 184.19\\
 & DH-Or-Mov & 1162.7 $\pm$ 325.95\\
 & DeepJanus & 42.97 $\pm$ 3.73 \\
 & DLFuzz & 111 $\pm$ 6.93\\
 
 \bottomrule

 \multirow{6}{*}{BeamNG} & DH-TurnCnt-StdSA & 139.8 $\pm$ 64.03  \\  
 & DH-TurnCnt-MinRad & 142.9 $\pm$ 59.54 \\
 & DH-MLP-MinRad & 161.3 $\pm$ 43.48 \\
 & DH-MLP-TurnCnt & 213 $\pm$ 87.63 \\
 & DH-MLP-StdSA & 183 $\pm$ 87.11 \\
 & DeepJanus & 5.2 $\pm$ 0.98 \\

 \bottomrule
 
\end{tabular}

\label{table:totalmisb}
\end{table}
\vskip -1em

\begin{tcolorbox}
\textbf{Summary}: \textit{\tool can find diverse failure-inducing inputs for all feature combinations. It can detect up to 10X more than the competitor in \bng.} 
\end{tcolorbox}

\subsection{RQ2: Search Exploration}

\begin{figure}
\centering
	\includegraphics[width=\columnwidth]{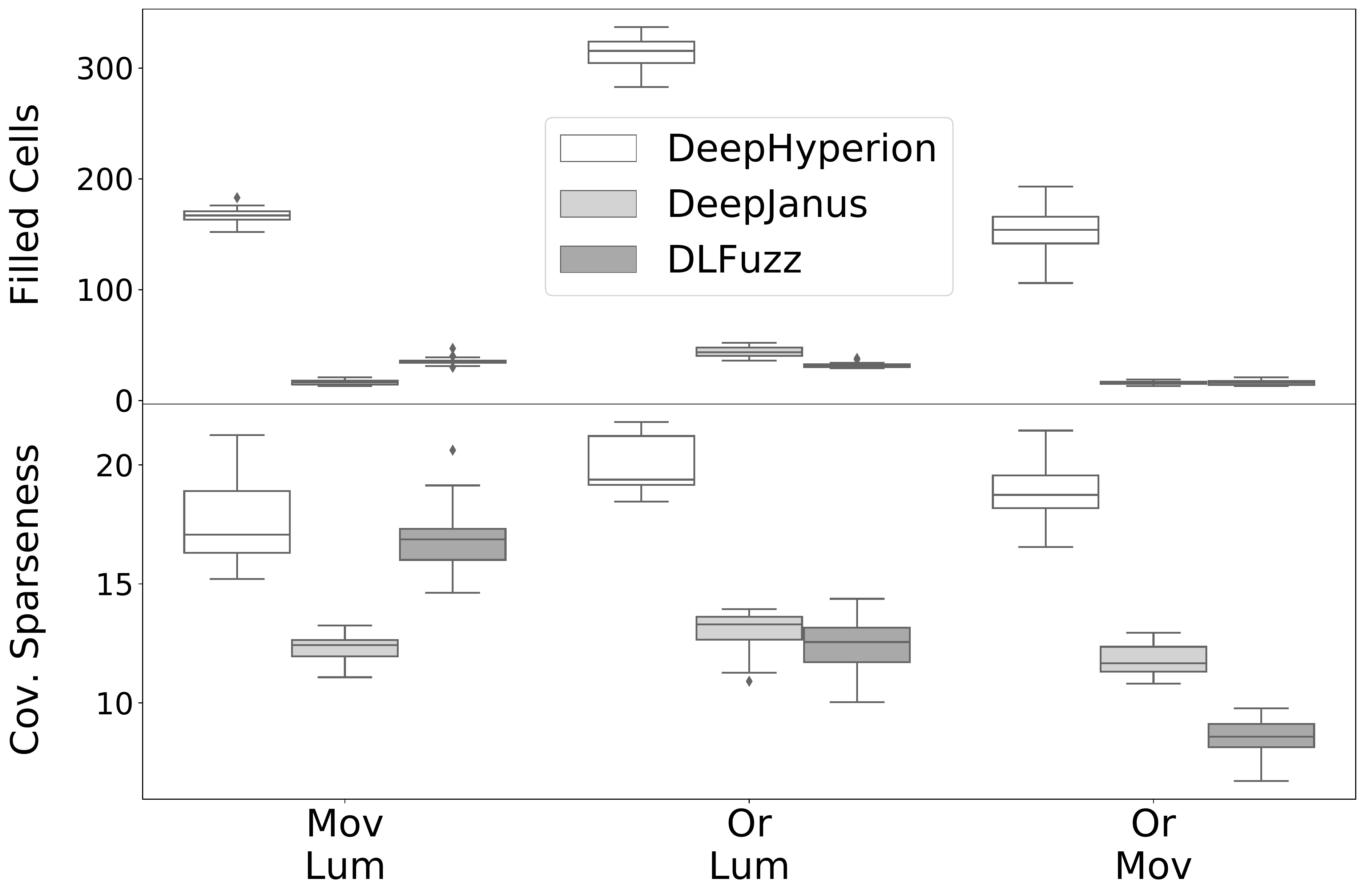}
        \caption{RQ2: Map cells filled by \tool, \djan and \dlf on \mnist}
        	\label{fig:rq2-MNIST}
\end{figure}

\autoref{fig:rq2-MNIST} shows the number and the sparseness of the filled cells in the maps produced by each tool for the \mnist system. 
\autoref{fig:rq2-MNIST} (top) shows that \tool covered all feature maps more extensively than the other tools (with large effect size and $p$-value always $<0.05$). Similarly to RQ1, the \textsc{Or-Lum} combination shows dramatically better results of \tool in comparison to the other tools, i.e. $>250$ additional cells filled by \tool. 

\autoref{fig:rq2-MNIST} (bottom) shows that \tool produced more sparse inputs for all three feature combinations. As regards the \textsc{Mov-Lum} combination, \tool has significantly better Coverage Sparseness than \djan and \dlf ($p$-value~$< 0.05$), with large effect size  vs \djan and small vs \dlf.

\begin{figure}
\centering
	\includegraphics[width=\columnwidth]{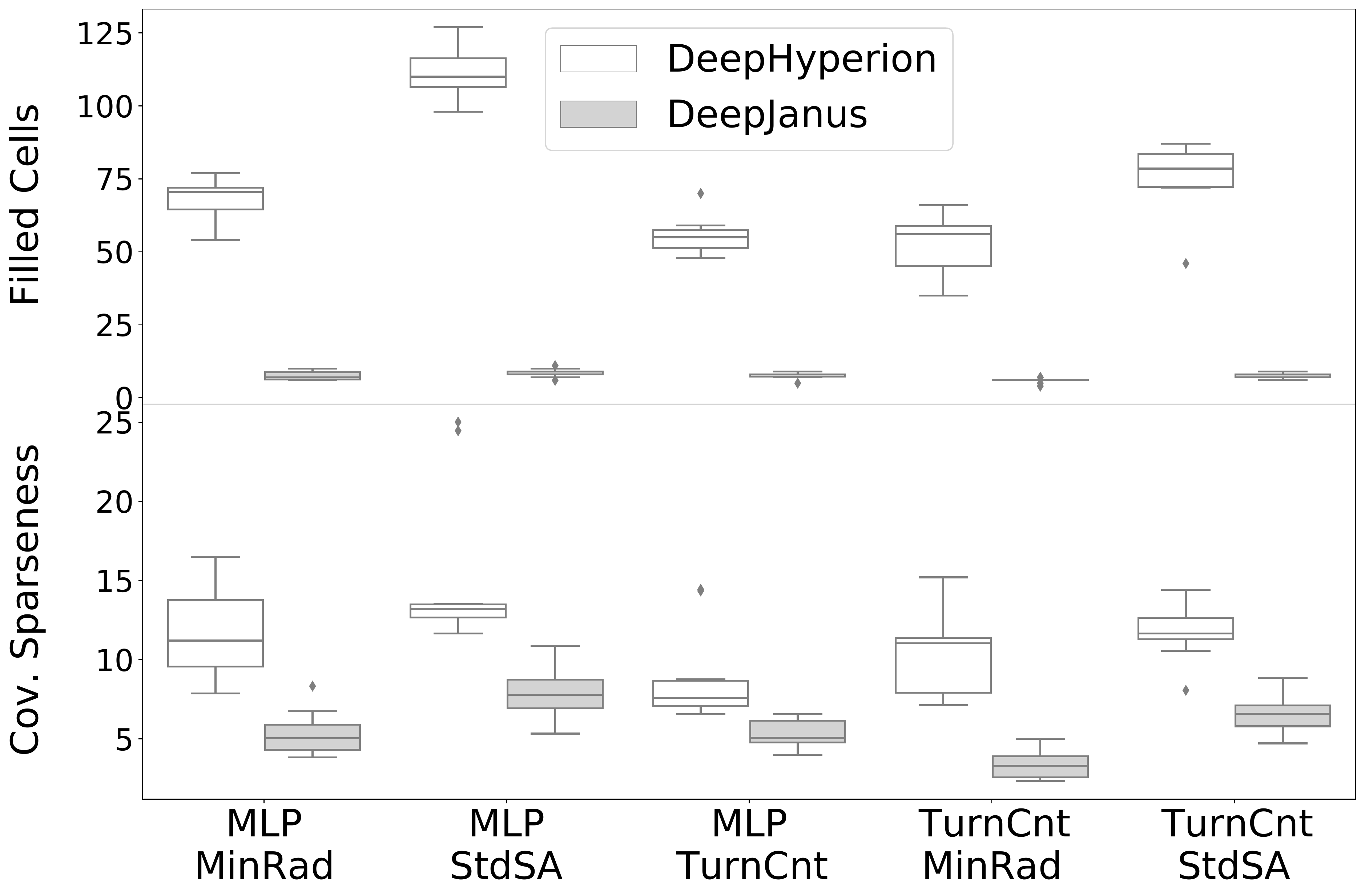}
        \caption{RQ2: Map cells filled by \tool and \djan on \bng}
	\label{fig:rq2-BeamNG}
\end{figure}

For the \bng system, \autoref{fig:rq2-BeamNG} reports Filled Cells and Coverage Sparseness achieved by the tools applied to the \bng system.
\autoref{fig:rq2-BeamNG} (top) confirms that \tool is particularly good in covering the map corresponding to the combination of behavioural features \textsc{MLP-StdSA} ($111.8$ filled cells on average). In comparison with \djan, it always filled significantly more cells (at least $47.7$ cells more).
\autoref{fig:rq2-BeamNG} (bottom) shows that \tool produced inputs that are always significantly sparser than \djan ($p$-values $<0.05$, large effect size).


\begin{tcolorbox}
\textbf{Summary}: \textit{\tool can always explore the feature space more extensively than the other tools (up to 8X for \mnist).} 
\end{tcolorbox}

\subsection{RQ3: Feature Discrimination}

\begin{figure*}[t!]
\centering
\includegraphics[width=\textwidth]{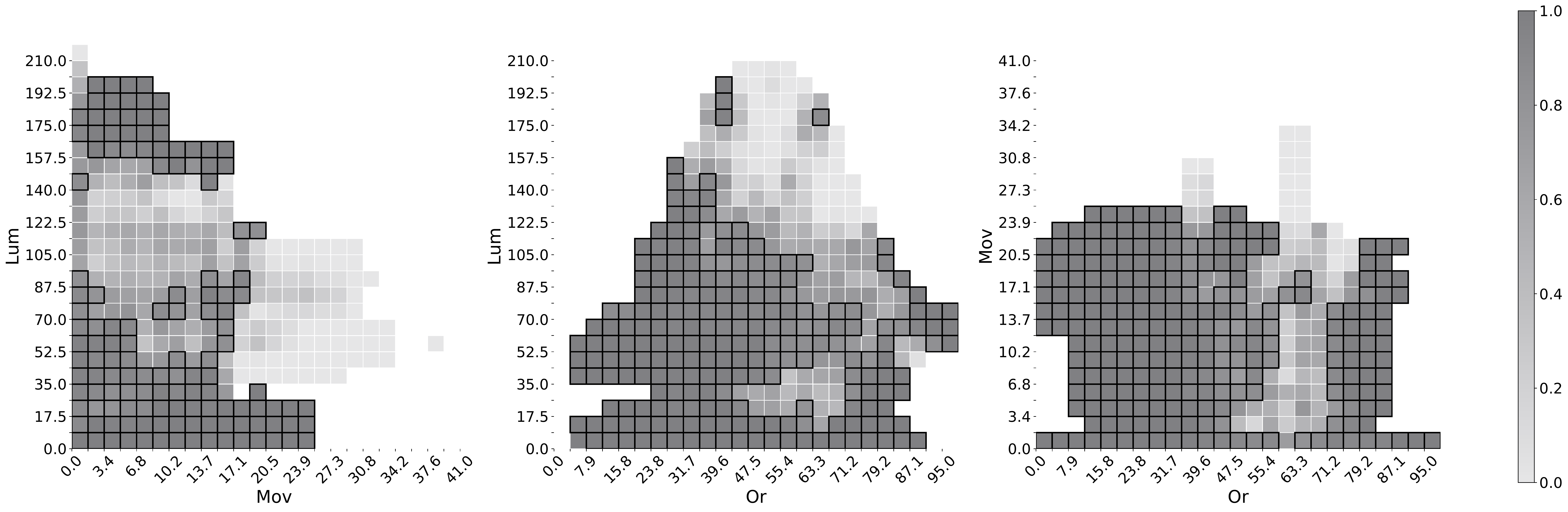}
\caption{RQ3: Probability maps and feature discrimination for \mnistts}
\label{fig:rq3-MNIST}
\includegraphics[width=\textwidth]{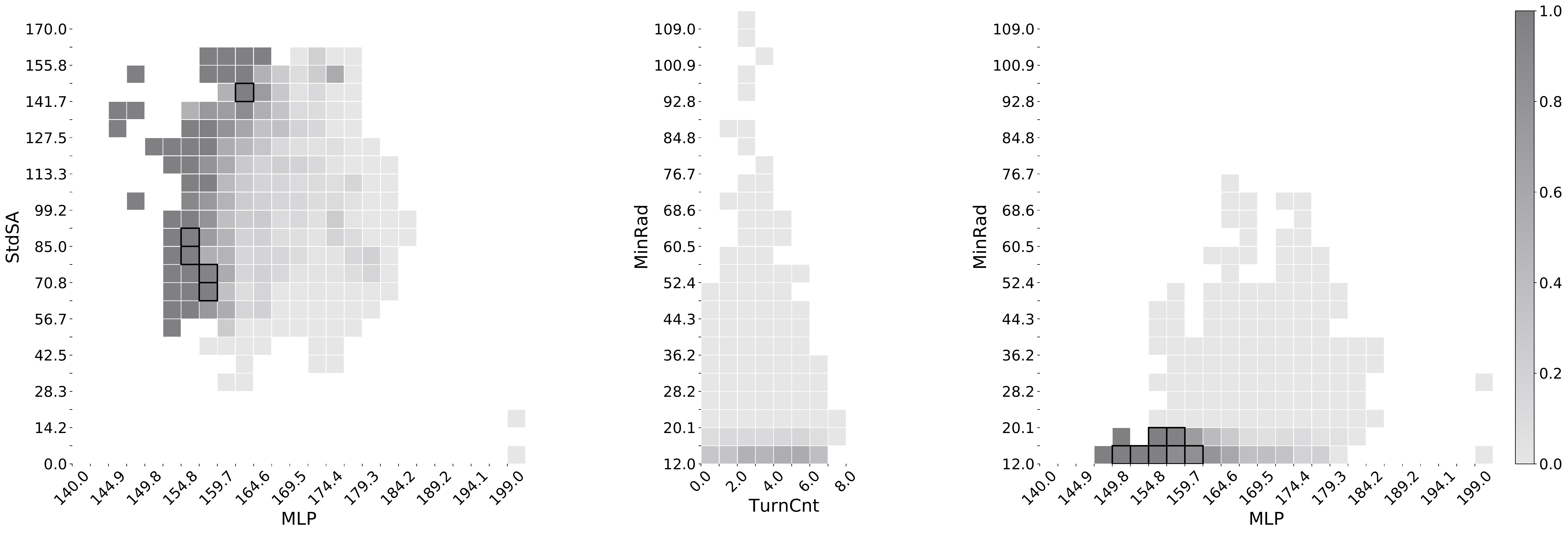}
\caption{RQ3: Probability maps and feature discrimination for \beamngts}
\label{fig:rq3-BeamNG}
\end{figure*}

In Figures~\ref{fig:rq3-MNIST} and~\ref{fig:rq3-BeamNG}, we report feature maps where the cell colour indicates the average Misbehaviour Probability (MP) across the runs for the corresponding feature combination (darker colour indicates higher probability). Blank cells are combinations that have never been explored by \tool. We highlight with a dark border the cells for which there is a MP > 0.8 and the lower bound of its confidence interval is > 0.65, which means that whenever an input is placed in these cells it is very likely to trigger a misbehaviour.

As regards \mnist, ~\autoref{fig:rq3-MNIST} shows that each feature map produced by \tool has multiple regions where the probability of failure is high. For instance,  ~\autoref{fig:rq3-MNIST} (center) suggests that left-oriented and thin digits are very likely to cause a classification failure. \tool can further help the user to interpret its results by showing the most representative inputs for each cell. As an example, in~\autoref{fig:fives} we show the actual inputs generated by \tool representing the map in ~\autoref{fig:rq3-MNIST} (left). We can see that bold are hard to recognise as ``5'' when part of the figure forms a circle, as they can be considered as ``6'' or ``9''. The bottom of the map shows that thin and discontinuous figures are hard to classify.

As regards BeamNG, ~\autoref{fig:rq3-BeamNG} shows that each of the maps corresponding to the combinations \textsc{MLP-StdSA} and \textsc{MLP-MinRad} has a clear region where the probability of failure is high. ~\autoref{fig:rq3-BeamNG} (left) suggests that the car is likely to go astray in roads that cause it to drive closer to the lane margins (lower \textsc{MLP}) and change often the steering angle direction (higher \textsc{StdSA}). ~\autoref{fig:rq3-BeamNG} (right) suggests that roads with at least a very sharp turn that cause the car to drive close to the lane margins are likely to cause a failure. The absence of high failure probability regions for the combination of the \textit{structural} features \textsc{TurnCnt-MinRad} (see ~\autoref{fig:rq3-BeamNG} (center)) may indicate that \textit{behavioural} features are more useful for characterising the conditions that trigger a misbehaviour.

\begin{tcolorbox}
\textbf{Summary}: \textit{For both subjects, \tool can detect well-characterised regions of the feature space that are likely to expose failures.} 
\end{tcolorbox}

\subsection{Threats to Validity}

\textbf{Construct Validity}:  \tool highly depends on map dimensions corresponding to measurable features. There is the risk that the selected features are not accurately quantified by the adopted metrics. To mitigate this threat, we (1) developed an empirical methodology to define the features of interest and the associated metrics, and (2) used metrics widely adopted in the literature. 

\noindent\textbf{External Validity}: The choice of subject DL systems is a possible threat to the \textit{external validity}. To mitigate this threat, we chose two diverse DL systems. One solves a classification problem, while the other  is a self-driving car software that solves a regression problem. However, further studies with a wider set of DL systems should be carried out to fully assess the generalisability of our findings.

\changed{\noindent\textbf{Conclusion Validity}: Random variations might have affected the results, given the highly stochastic nature of DL systems. To mitigate this threat, we ran each experiment multiple times and statistical tests to assess the significance of our results, according to the guidelines for comparing randomised test generation algorithms proposed by Arcuri and Briand~\cite{Arcuri14}.}

\section{Related Work}
\label{related}
DL systems' quality has been mainly assessed by generating new inputs that expose misbehaviours~\cite{PeiCYJ17,GuoJZCS18,TianPSB18,Ma-ASE-2018,GambiMF19} and by proposing novel adequacy criteria that guide the input generation process~\cite{PeiCYJ17,XieISSTA19,KimFY19}. To the best of our knowledge, no technique aims at covering the feature space of DL systems and few works~\cite{AbdessalemNBS18,RiccioFSE20} make use of interpretable properties for test generation.

\subsection {Input Generation and Adequacy} 

DeepXplore~\cite{PeiCYJ17} is a testing technique to detect behaviour inconsistencies among different DNNs.  It is guided by neuron coverage, i.e., the percentage of neurons whose activation level is above a certain threshold. DLFuzz \cite{GuoJZCS18} is also a test input generator guided by neuron coverage. DeepTest~\cite{TianPSB18} maximises the neuron coverage of a DNN-based steering angle predictor by applying different image transformations to images captured by the on-board camera of an autonomous car. DeepGauge~\cite{Ma-ASE-2018} uses a set of coverage criteria that extend neuron coverage by taking the distribution of training data into consideration. DeepCT~\cite{LeiJXLLLZ19} uses a set of combinatorial testing criteria for DNNs based on the interactions between neurons. DeepHunter~\cite{XieISSTA19} leverages multiple coverage criteria originally proposed by Ma et al.~\cite{Ma-ASE-2018} as feedback to guide test generation. DeepSmartFuzzer~\cite{SametFS19} uses Monte Carlo Tree Search (MCTS) to exploit the coverage increase patterns.  The degree of ``surprise'' of an input was measured  by means of the two metrics proposed by Kim et al.~\cite{KimFY19}, associated with a \textit{surprise adequacy} coverage criterion.

The above test input generation techniques focus on generating adversarial inputs by adding perturbations to the original inputs. 
Other input generators~\cite{RiccioFSE20,GambiMF19,AbdessalemNBS18} are instead based on the manipulation of a model of the inputs, which ensure more control on the validity and realism of the generated inputs, going beyond adversarial attacks that expose security vulnerabilities. 
For instance, DeepJanus~\cite{RiccioFSE20} manipulates the way-points that define the shape of a road within a self-driving car simulator. AsFault \cite{GambiMF19} is a model-based approach that applies a search-based algorithm to test the lane-keeping system of self-driving cars.
\tool belongs to this category of test generators.

All  test generators mentioned above aim at maximising some internal adequacy metric, such as neuron or surprise coverage, or at exposing misbehaviours. None of them considers the value combinations of interpretable features of the DL system under test as the target of test generation. \tool is the first tool to provide developers with a map of such  features, where the automatically generated inputs, as well as the exposed misbehaviours, are positioned and can be interpreted. Hence, existing test generators might completely ignore parts of a feature map or might expose only misbehaviours that belong to a narrow map region.

\subsection{Structural and Behavioural Properties} 

NSGAII-DT~\cite{AbdessalemNBS18} is a model-based approach for testing vision-based control systems. This approach builds on  evolutionary multi-objec\-ti\-ve algorithms and uses decision trees to guide the generation of new test scenarios within the multidimensional space of the model parameters. 
Decision trees are used to identify the critical regions of the input space, i.e., the combinations of model parameter values that are more likely to cause misbehaviours. While decision trees provide interpretable information to developers as \tool does with its feature maps, the variables that appear in decision nodes are limited to the control parameters of the input model, which might not be fully representative of all relevant behavioural features of the system under test. Moreover, decision trees are used to focus the search on critical scenarios (collisions or near-collision at high speed with pedestrians), so as to increase the search efficiency, while \tool aims at covering the feature map at large, so as to ensure that as many regions as possible are tested and that regions with misbehaviours are not left untested.

DeepJanus~\cite{RiccioFSE20} characterises the quality of a DL system as its frontier of behaviours, i.e., pairs of similar inputs that trigger different (expected vs failing) behaviours of the system. The output of DeepJanus provides users with a a set of  system's frontier inputs, but it does not explicitly characterise them  based on structural or behavioural features. Instead, \tool's maps allow developers to interpret the inputs that trigger a misbehaviour in terms of their feature values.

The properties we use as feature dimensions are identified by experts during the open coding step of our empirical methodology. In the literature, weak supervision approaches~\cite{weaksup_survey}, e.g., the Data Programming paradigm~\cite{Ratner16}, also exploit domain-experts' knowledge to create and assign output labels to the training set elements. Unlike these approaches, our open coding identifies input features that can be quantified by metrics, without considering their relationship with the network's output.
\section{Conclusions and Future Work}
\label{conclusions}

\tool provides a unique characterisation of a DL system's quality through an interpretable map which represents the highest-performing (i.e., misbehaving or closest to misbehaving) inputs in the space of the relevant, domain-specific features.

Our empirical study shows that \tool is more effective than state-of-the-art DL testing tools in generating failure-inducing inputs associated with highly diverse features. In the reverse direction, we showed that \tool is useful to detect the feature combinations that are most likely to induce a system misbehaviour. In our future work, we plan to generalise our results to a wider sample of DL systems, including industrial ones.

\begin{acks}
This work was partially supported by the H2020 project PRECRIME, funded under the ERC Advanced Grant 2017 Program (ERC Grant Agreement n. 787703). The driving simulator has been provided by BeamNG GmbH. 
\end{acks}

\bibliographystyle{ACM-Reference-Format}
\balance
\bibliography{biblio}

\end{document}